%% file: main.tex
\documentclass[a4paper,runningheads]{llncs}

\usepackage{ifthen}

\newboolean{tosubmit}
\setboolean{tosubmit}{true}

\newboolean{showhelpers}
\setboolean{showhelpers}{true}

\usepackage[british]{babel}
\usepackage[T1]{fontenc}
\usepackage[utf8x]{inputenc}
\usepackage[font=small,labelfont=bf]{caption}

\usepackage{montserrat}
\usepackage{lmodern}

\DeclareFontFamily{U}{mathx}{\hyphenchar\font45}
\DeclareFontShape{U}{mathx}{m}{n}{
      <5> <6> <7> <8> <9> <10>
      <10.95> <12> <14.4> <17.28> <20.74> <24.88>
      mathx10
      }{}
\DeclareSymbolFont{mathx}{U}{mathx}{m}{n}
\DeclareFontSubstitution{U}{mathx}{m}{n}
\DeclareMathAccent{\widecheck}{0}{mathx}{"71}
\usepackage{amsmath,
            amsthm,
            amssymb,
            amsfonts}
\usepackage{amsbsy}    
\usepackage[mathscr]{eucal}                      
\usepackage{cutwin}  
\usepackage{stmaryrd}
\usepackage{mathtools}  
\usepackage{calrsfs}
\usepackage{bm}
\DeclareMathAlphabet{\pazocal}{OMS}{zplm}{m}{n}
\usepackage{subcaption,graphicx,wrapfig}
\usepackage{xfrac,xspace,xifthen}
\usepackage{twoopt}  
\usepackage{colortbl}
\usepackage[inline, shortlabels]{enumitem}
\usepackage{array,booktabs,multirow} 
\usepackage[normalem]{ulem}
\usepackage[symbol*,hang,marginal]{footmisc}  
\usepackage{listings,algorithm,algorithmicx, algpseudocode}
\usepackage{tikz,tikzsymbols,pgfplots}
\usepackage{relsize}
\usepackage{breakcites}
\usepackage{lipsum}
\usepackage{longtable}
\usepackage{lscape}
\usepackage{tabularx, xtab, booktabs}
\allowdisplaybreaks
\usepackage{comment}
\usepackage{soul}

\graphicspath{{imgs/}{figures/}}  


\makeatletter
\RequirePackage[%
  bookmarks,
  unicode,
  colorlinks=true,
  breaklinks=true]%
  {hyperref}%
    \def\@citecolor{blue}%
    \def\@urlcolor{blue}%
    \def\@linkcolor{blue}%

\def\orcidID#1{\href{http://orcid.org/#1}{\protect\raisebox{-1.25pt}{\protect\includegraphics[height=8pt]{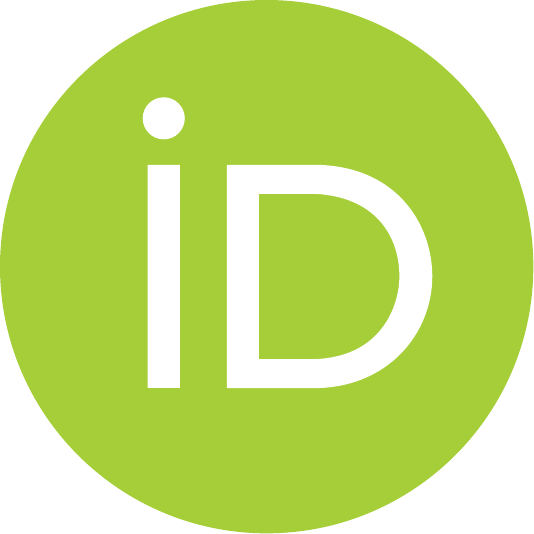}}}}
\makeatother
\usepackage[nameinlink]{cleveref}
\usepackage[outline]{contour}
\contourlength{0.115mm} 
\definecolor{lightblue}{RGB}{231,255,255}
\definecolor{lightred}{RGB}{255,224,224}
\definecolor{lightgreen}{RGB}{224,255,224}
\definecolor{lightyellow}{RGB}{255,255,224}
\definecolor{lightpurple}{RGB}{255,224,255}
\definecolor{darkerred}{RGB}{64,0,0}
\definecolor{darkred}{RGB}{128,0,0}
\definecolor{darkblue}{RGB}{0,0,128}
\definecolor{darkgreen}{RGB}{0,128,0}
\definecolor{darkpurple}{RGB}{128,0,128}
\definecolor{black}{RGB}{0,0,0}
\definecolor{white}{RGB}{255,255,255}
\definecolor{lightgray}{RGB}{240, 240, 245}

\definecolor{pathogens}{RGB}{255,133,255}
\definecolor{transmission}{RGB}{102,255,255}
\definecolor{hosts}{RGB}{255,179,102}
\definecolor{infected}{RGB}{204,153,255}
\definecolor{physical}{RGB}{153,255,51}
\definecolor{status}{RGB}{255,102,102}
\definecolor{error}{RGB}{217,217,217}
\definecolor{primodfail}{RGB}{102,178,255}
\definecolor{defectspipe}{RGB}{255, 255, 51}
\definecolor{interference}{RGB}{30, 250, 162}
\definecolor{geo}{RGB}{235, 174, 174}
\definecolor{design}{RGB}{176, 255, 41}
\definecolor{operation}{RGB}{212, 225, 245}
\definecolor{corrosion}{RGB}{226, 204, 255}
\definecolor{corrosionThin}{RGB}{255, 194, 102}
\definecolor{corrosionMedium}{RGB}{179, 255, 255}
\definecolor{dos}{RGB}{194, 196, 255}
\definecolor{infogather}{RGB}{255, 102, 102}
\definecolor{datatamper}{RGB}{204, 153, 255}
\definecolor{accessgs}{RGB}{102, 255, 102}
\definecolor{killcomms}{RGB}{255, 246, 120}
\definecolor{oars}{RGB}{0, 110, 255}

\usetikzlibrary{arrows,arrows.meta,backgrounds,calc,shapes,pgfplots.groupplots,positioning,automata,plotmarks,shapes.gates.logic.US,shapes.gates.logic.IEC}
\tikzstyle{every picture}+=[remember picture, overlay, anchor=center]
\tikzstyle{every node}=[initial text=]
\tikzstyle{every picture}=[>=stealth]
\tikzstyle{BAS}=[ellipse, draw, fill=normalfill, minimum width=7mm, inner sep=1]

\lstdefinelanguage{Galileo}{
  keywords={},  
  keywords=[2]{toplevel,and,or,pand,csp,wsp,hsp,fdep,seq,
               2of3,2of4,3of4,2of5,3of5,4of5,2of6,3of6,4of6,5of6},
  keywords=[3]{prob,lambda,dorm,phases,interval,res,repair},
  keywords=[4]{insp1,insp2,insp3,insp4,insp5,1insp1,1insp2,2insp2,
               1insp3,2insp3,3insp3,1insp4,2insp4,3insp4,4insp4,
               1insp5,2insp5,3insp5,4insp5,5insp5},  
  otherkeywords={=},
  morecomment=**[l]{//},      
  morecomment=**[s]{/*}{*/},
  morestring=[b]",
  moredelim=**[is][\color{codenumeric}]{^}{^},     
  moredelim=**[is][\color{codesyncaction}]{~}{~},  
  moredelim=**[is][\slshape]{__}{__},              
}
\Crefname{equation}{eq.}{eqs.}
\crefname{equation}{equation}{equations}
\Crefname{figure}{Fig.}{Figs.}
\crefname{figure}{figure}{figures}
\crefname{tabular}{table}{tables}
\Crefname{definition}{Def.}{Defs.}
\crefname{definition}{definition}{definitions}
\Crefname{proposition}{Prop.}{Props.}
\crefname{proposition}{proposition}{propositions}
\Crefname{section}{Sec.}{Secs.}
\crefname{section}{section}{sections}
\crefname{algorithm}{Algo.}{algorithms}
\crefname{listing}{code}{codes}
\Crefname{example}{Ex.}{Examples}
\crefname{example}{ex.}{examples}

\DefineFNsymbols*{lamport}[math]{
  {\hspace{1pt}\bigstar\hspace{1pt}}
  {\hspace{2pt}\dagger\hspace{1pt}}
  {\hspace{1pt}\ddagger\hspace{.2pt}}
  {\hspace{1pt}\ast\hspace{.2pt}}
  {\hspace{1pt}\S\hspace{.2pt}}
  {\hspace{1pt}\P\hspace{.2pt}}
  {\hspace{1pt}\|\hspace{.2pt}}
  {\hspace{1pt}\dagger\dagger\hspace{.2pt}}
  {\hspace{1pt}\ddagger\ddagger\hspace{.2pt}}
  {\hspace{1pt}\ast\ast\hspace{.2pt}}
}
\setfnsymbol{lamport}

\ifthenelse{\boolean{showhelpers}}{%
  \usepackage[pagewise]{lineno}  
}{}

\ifthenelse{\boolean{tosubmit}}{}{ 
  \usepackage{draftwatermark}  
}

\makeatletter
\renewcommand{\paragraph}{\@startsection{paragraph}{5}{0em}%
  {.7ex plus .2ex minus .1ex}%
  {-.5em}%
  {\bfseries}}
\makeatother

\makeatletter
\def\THICKhrulefill{\leavevmode \leaders \hrule height 5pt\hfill \kern \z@}
\makeatother
\newcommand{\colorpar}[3]{\colorbox{#1}{\parbox{#2}{#3}}}
\newcommand{\marginremark}[3]{%
  \ifthenelse{\boolean{tosubmit}}{}{
	\marginpar{\colorpar{#2}{.65\linewidth}{\color{#1}#3}}
}}
\newcommand{\highlightedremark}[4]{%
  \ifthenelse{\boolean{tosubmit}}{}{
	\begin{center}\fcolorbox{#1}{#2}{%
	\begin{minipage}{.98\linewidth}\color{#1}%
	\textbf{\THICKhrulefill[ #3 ]\THICKhrulefill}%
	\par\noindent#4\end{minipage}}\end{center}%
}}

\newcommand{\rmkStefano}[1]{\marginremark{darkblue}{darkblue!10}{\tiny{[SMN]~ #1}}}

\newcommand{\todo}[1][]{%
  \ifthenelse{\boolean{tosubmit}}{}{
  \textbf{\sffamily\textcolor{Red}{TODO:} #1}%
  \marginpar{\textsf{\color{red}\bfseries TODO}}}}

\renewcommand{\top}{\mathtt{{1}}}
\renewcommand{\bot}{\mathtt{{0}}}

\renewcommand{\vec}{\mathaccent "017E\relax}  
\newcommand\restr[2]{{
  \left.\kern-\nulldelimiterspace #1 \vphantom{\big|} \right|_{#2}}}
\newcommand{\BB}{\ensuremath{\mathbb{B}}\xspace}  
\newcommand{\from}{\colon}

\newcommand{\acronym}[1]{\ensuremath{\text{\uppercase{#1}}}\xspace}
\newcommand{\mathobject}[1]{\ensuremath{\text{\scalebox{.92}{$#1$}}}}

\newcommand{\Vars}{\ensuremath{\mathit{Vars}}\xspace}
\DeclareMathOperator{\bddOp}{\mathobject{B}}

\newcommand{\BDDlOne}{\ensuremath{\bddOp_{\odg}^{\lOne}}\xspace}
\makeatletter  
\newcommand{\precneq}{\mathrel{\text{\prec@eq}}}
\newcommand{\prec@eq}{%
  \oalign{%
    \hidewidth$\m@th\prec$\hidewidth\cr
    \noalign{\nointerlineskip\kern1ex}%
    $\m@th\smash{\raisebox{0.65ex}{\rotatebox{90}{%
      \scalebox{1.1}[-1.1]{$\nshortmid$}}}}$\cr
    \noalign{\nointerlineskip\kern-.5ex}%
}}
\makeatother

\DeclareMathOperator*{\argmin}{argmin}
\DeclareMathOperator*{\argmax}{argmax}
\newcommand{\AT}{\acronym{at}}               
\newcommand{\ATs}{\acronym{at}{s}\xspace}

\newcommand{\FT}{\acronym{ft}}               
\newcommand{\FTs}{\acronym{ft}{s}\xspace}
\newcommand{\DT}{\acronym{dt}}               
\newcommand{\DTs}{\acronym{dt}{s}\xspace}

\newcommand{\BDD}{\acronym{bdd}}             
\newcommand{\BDDs}{\acronym{bdd}{s}\xspace}

\newcommand{\BU}[1][]{\ensuremath{\mathtt{BU%
  \ifthenelse{\isempty{#1}}{}{_{\mkern1mu#1}}}}\xspace}

\newcommand{\nodeType}[1]{\ensuremath{\mathtt{#1}}\xspace}
\newcommand{\tBAS}{\nodeType{BAS}}
\newcommand{\tBE}{\nodeType{BE}}
\newcommand{\tOR}{\nodeType{OR}}
\newcommand{\tAND}{\nodeType{AND}}

\DeclareMathOperator{\chOp}{\mathit{ch}}       

\newcommand{\child}[1]{\ensuremath{\chOp({#1})}\xspace}

\DeclareMathOperator{\sfun}{\mathit{f}}        
\DeclareMathOperator{\xfun}{\mathit{f}^\circ}  
\newcommand{\sfunT}[1][\!\,\T]{\ensuremath{\sfun_{\!#1}}\xspace}
\newcommand{\xfunT}[1][\!\!\!\!\tree]{\ensuremath{\xfun_{\!#1}}\xspace}

\newcommand{\xfunA}[1][\!\!\!\A]{\ensuremath{\xfun_{\!#1}}\xspace}

\newcommand{\xfunF}[1][\!\!\!\F]{\ensuremath{\xfun_{\!#1}}\xspace}
\newcommand{\minSat}[1]{\ensuremath{
  \llbracket{#1}\rrbracket_{\scriptscriptstyle \lOneModel}}\xspace}

\newcommand{\before}[1][]{\mathbin{%
  \ifthenelse{\isempty{#1}}{%
    \tikz[baseline=-.6ex]{\draw[->,thin,x=1ex,y=1ex](0,0)--(1.7,0);}%
  }{%
    \tikz[baseline=-.48ex]{\draw[->,thin,x=1ex,y=1ex](0,0)--(1.7,0);%
    \node[x=1ex,y=1ex,inner sep=0pt]at(.75,.75){$\scriptscriptstyle{#1}$};}}}
}
\newcommand{\op}{\ensuremath{op}}
\newcommand{\conf}{\ensuremath{\vecb_{\!O}}}
\newcommand{\lOneModel}{\ensuremath{\mathcal{M}}}
\newcommand{\lTwoModel}{\ensuremath{\mathsf{M}}}
\newcommand{\scenario}{\ensuremath{\vecb}}
\newcommand{\riskscenario}{\scenario_{\!\!R}}
\newcommand{\attack}[1][\scenario_{\!\!A}]{\mathobject{#1}\xspace}
\newcommand{\fault}[1][\scenario_{\!F}]{\mathobject{#1}\xspace}

\newcommand{\allConfig}{\ensuremath{\mathcal{C}}\xspace}
\newcommand{\allConfigPrime}{\ensuremath{\mathcal{C}_{[op\mapsto bool]}}\xspace}
\newcommand{\allScenarios}{\ensuremath{\mathcal{S}}\xspace}

\newcommandtwoopt{\poset}[2][\attack][\prec]{%
  \ensuremath{\langle{#1},{#2}\rangle}\xspace}
\newcommandtwoopt{\Hasse}[2][\attack][\prec]{%
  \ensuremath{\mathobject{H}_{\mkern-2mu#1}^{#2}}\xspace}

\DeclareMathOperator{\attrOp}{\alpha}

\newcommand{\attr}[1]{\ensuremath{\attrOp(#1)}\xspace}    

\newcommand{\problOne}[1]{\ensuremath{\rho(#1)_{A, F}}} 

\newcommand{\OR}{\acronym{or}}               
\newcommand{\AND}{\acronym{and}}             
\newcommand{\BASs}{\acronym{bas}{es}\xspace}
\newcommand{\BASes}{\BASs}
\newcommand{\T}[1][]{\ensuremath{\mathobject{T}\ifthenelse{\isempty{#1}}{}{_{\hspace{-2pt}#1}}}\xspace}
\newcommand{\A}[1][]{\ensuremath{\mathobject{A}\ifthenelse{\isempty{#1}}{}{_{\hspace{-2pt}#1}}}\xspace}
\newcommand{\F}[1][]{\ensuremath{\mathobject{F}\ifthenelse{\isempty{#1}}{}{_{\hspace{-2pt}#1}}}\xspace}

\newcommand{\bddT}[1][\odg]{\ensuremath{\bddOp_{#1}}\xspace}

\newcommand{\alphaOdg}{\alpha_{\!\odg}}

\renewcommand{\top}{\mathtt{{1}}}
\renewcommand{\bot}{\mathtt{{0}}}

\newcommand{\Child}{\ensuremath{\mathit{ch}}}

\newcommand{\lOne}{\ensuremath{\phi}}
\newcommand{\lTwo}{\ensuremath{\psi}}

\newcommand{\lThree}{\ensuremath{\xi}}

\newcommand{\lTwoThree}{\ensuremath{\theta}}

\newcommand{\Trans}{\mathfrak{f}}

\newcommand{\valFun}{\ensuremath{\mathsf{Val}_\mathcal{C}}}
\newcommand{\partRisk}{\ensuremath{\mathsf{objRiskVal}}}

\newcommand{\tree}{\textit{T}\xspace}
\newcommand{\ODG}{\textit{DOG}\xspace}
\newcommand{\ODGs}{\acronym{DOG}{s}\xspace}
\newcommand{\odg}{\textit{G}\xspace}
\newcommand{\OaR}{OaR\xspace}
\newcommand{\TLE}{\acronym{tle}}

\newcommand{\MCSs}{\acronym{mcs}{s}\xspace}
\newcommand{\MRSs}{\acronym{mrs}{s}\xspace}

\newcommand{\OaRs}{{OaR}{s}\xspace}

\newcommand{\BE}{\acronym{be}}
\newcommand{\BEs}{\acronym{be}{s}\xspace}

\newcommand{\IEs}{\acronym{ie}{s}\xspace}

\newcommand{\LEAF}{\acronym{leaf}}            
\newcommand{\LEAFs}{\acronym{leave}{s}\xspace}

\newcommand{\tIE}{\nodeType{IE}}
\newcommand{\tLEAF}{\nodeType{LEAF}}

\newcommand{\opMRS}{\mathrm{MRS}}
\newcommand{\opMSS}{\mathrm{MSS}}

\newcommand{\tV}{\Vars}
\newcommand{\tVprime}{\tV'}

\newcommand{\OurFrame}{\ensuremath{\mathsf{WATCHDOG}}}
\newcommand{\OurLogic}{\ensuremath{\mathsf{DOGLog}}}
\newcommand{\OurDSL}{\ensuremath{\mathsf{DOGLang}}}
\newcommand{\vecb}{\ensuremath{\vec{b\,}}}

\newcommand{\NA}{\ensuremath{N_{\!A}}}
\newcommand{\NF}{\ensuremath{N_{\!F}}}
\newcommand{\NO}{\ensuremath{N_{\!O}}}

\algblockdefx[Foreach]{Foreach}{EndForeach}[1]{\textbf{for each} #1 \textbf{do}}{\textbf{end for}}

\hypersetup{  
  pdftitle = {WATCHDOG: an ontology-aWare risk AssessmenT approaCH via object-oriented DisruptiOn Graphs}
}

\pgfplotsset{compat=1.16}
\begin{document}
\title{%
	 WATCHDOG: an ontology-aWare risk AssessmenT approaCH via object-oriented DisruptiOn Graphs\thanks{%
		This work was partially funded by the NWO grant NWA.1160.18.238 (PrimaVera), the EU's Horizon 2020 Marie Curie grant No 101008233, ERC Consolidator and Proof of Concept  Grants 864075 (\textit{CAESAR}) and 101187945 (\textit{RUBICON}).}%
	}
\titlerunning{WATCHDOG: ontology-aware risk assessment via ODGs}
%

\author{%
    Stefano~M.~Nicoletti\inst{1}\orcidID{0000-0001-5522-4798}
	\and
    E.~Moritz~Hahn\inst{1}\orcidID{0000-0002-9348-7684}
	\and   
    Mattia~Fumagalli\inst{2}\orcidID{0000-0003-3385-4769}
    \and
    Giancarlo~Guizzardi\inst{1}\orcidID{0000-0002-3452-553X}
    \and
	Mari\"elle~Stoelinga\inst{1,3}\orcidID{0000-0001-6793-8165}
}
\authorrunning{S.M.~Nicoletti, E.M.~Hahn, M.~Fumagalli, G.~Guizzardi and M.~Stoelinga}

\institute{%
	University of Twente, Enschede, the Netherlands
	\and
    Free University of Bozen-Bolzano, Bozen, Italy
    \and
	Radboud University, Nijmegen, the Netherlands
    \email{\{s.m.nicoletti,e.m.hahn,g.guizzardi,m.i.a.stoelinga\}@utwente.nl, mattia.fumagalli@unibz.it}
\vspace{-10mm}
}
\maketitle
\setcounter{footnote}{1} 
\renewcommand{\thelstlisting}{\arabic{lstlisting}}  

%
%
\begin{abstract}
    When considering risky events or actions, we must not downplay the role of involved objects: a charged battery in our phone averts the risk of being stranded in the desert after a flat tyre, and a functional firewall mitigates the risk of a hacker intruding the network. The \textit{Common Ontology of Value and Risk} (COVER) highlights how the role of objects and their relationships remains pivotal to performing transparent, complete and accountable risk assessment. In this paper, we operationalize some of the notions proposed by COVER -- such as \textit{parthood} between objects and \textit{participation} of objects in events/actions -- by presenting a new framework for risk assessment: WATCHDOG. WATCHDOG enriches the expressivity of vetted formal models for risk -- i.e., \textit{fault trees} and \textit{attack trees} -- by bridging the disciplines of \textit{ontology} and \textit{formal methods} into an ontology-aware formal framework composed by a more expressive modelling formalism, \textit{Object-Oriented Disruption Graphs} (DOGs), logic (DOGLog) and an intermediate query language (DOGLang). With these, WATCHDOG allows risk assessors to pose questions about \textit{disruption propagation}, \textit{disruption likelihood} and \textit{risk levels}, keeping the fundamental role of objects at risk always in sight.
   \vspace{0.5mm}
\end{abstract}
\input{Text_Body}


\bibliographystyle{splncs04}
\bibliography{main-compact}


\end{document}

%% file: Text_Body.tex
\section{Introduction}
\label{sec:Intro}

Risk assessment is a key activity to identify, analyze and prioritize the risk in a system, and come up with \mbox{(cost-)}effective countermeasures \cite{zio2018future}. This is true when considering \textit{safety} (i.e., the absence of risk connected with unintentional malfunctions) and \textit{security} (i.e., the absence of risk linked with intentional attacks) \cite{stoelinga2021marriage}. To perform transparent, complete and accountable risk assessment, it is fundamental to explicitly account for the role objects play in \textit{Events} (including \textit{Actions}) in which they participate, and for how their status affects safety and security interplay: a door being locked causes the impossible escape event in case of fire but simultaneously stops the action of a burglar entering your house \cite{kriaa2014safety, Nicoletti2023Model, sun2009addressing}. Formalisms widely employed in industry and academia to conduct risk assessment -- such as \textit{fault trees} \cite{RS15b} and \textit{attack trees} \cite{Sch99} -- are not equipped to explicitly reason about objects. Without an explicit representation of objects at risk, it is impossible to evaluate their role in risk scenarios and to correctly evaluate the overall risk imposed on these objects: e.g., what is the risk level my car is subject to, considering that both its tyres can break (safety-related event) and its onboard computer can be hacked (security-related action)?


Our objective here is to provide an ontology-grounded formal approach for object-based risk representation and reasoning by combining and extending standard formalisms for safety (\textit{fault trees}) and security (\textit{attack trees}). To address this lack of expressivity, two promising fields must be taken into account: \textit{ontologies for risk} and \textit{model-based risk assessment}. On the one hand, risk ontologies -- like the \textit{Common Ontology of Value and Risk} (COVER) \cite{sales2018common} -- excel in providing a structured ground for reasoning about a specific domain of knowledge, transparently and explicitly laying out key concepts and relationships needed to reason about risk. While excellent for conceptualization and transparency, ontologies are however not designed to enable quantitative and applied risk evaluations. 
On the other hand, specific model-based technologies from the field of formal methods -- like \textit{fault trees} (\FTs) and \textit{attack trees} (\ATs) -- excel in providing applicable, tried and tested instruments for rigorous and quantitative risk assessment. These methods, however, sometimes rely on opaque conceptual assumptions and, in particular, do not offer the expressivity needed to explicitly reason about objects at risk. With $\OurFrame$ we propose a risk assessment framework that enriches and extends the expressivity of vetted model-based technologies -- such as \FTs and \ATs\ -- while grounding them in the conceptual clarity of COVER.

Fundamental elements highlighted by the COVER ontological framework \cite{sales2018common} -- such as the \textit{participation} of a given object in a risk-related \textit{action}/\textit{event} or the \textit{parthood} relationship between different objects -- are not expressible in classical risk assessment formalisms, such as \FTs and \ATs. We address this gap and operationalize these concepts by presenting \textit{object-oriented DisruptiOn Graphs} (DOGs), a new formalism that extends the strengths of classical \FT- and \AT-based risk analysis accounting for the role of objects at risk (\OaRs) in \textit{disruption propagation}, \textit{likelihood} and \textit{risk calculation}\footnote{We adopt these notions from previous work. For a more in-depth discussion on \textit{risk}, \textit{disruption}, \textit{propagation}, and their relation with \textit{risk propagation}, see \cite{fumagalli2023semantics, stoelinga2021marriage}.}. Morever, to perform transparent decision-making w.r.t. safety and security of systems, practitioners need the ability to analyse their models in a meaningful and thorough way. To cater for this need, we present $\OurLogic$ -- a logic to formally query \ODGs -- and $\OurDSL$, an intermediate domain-specific language to ease the querying process. With $\OurLogic$ and $\OurDSL$ practitioners can query \ODGs to learn meaningful information about systems \textit{disruptions} and \textit{risk levels}.
One could ask, for example: Given that one of my tyres breaks, is my entire field trip compromised? Is the probability of an attacker compromising the network larger/smaller than \textit{p}? Given an object at risk (e.g., my laptop), what is the most risky security Action/safety Event in which it participates? What is the maximum risk level imposed on my laptop, given all the Actions/Events in which it participates?
Finally, we showcase property specification in $\OurLogic$ and $\OurDSL$ on an \ODG model for a variant of small but well-known and representative example from safety-security literature, modelling safety and security risks on a household given the status of a door lock \cite{kriaa2014safety, Nicoletti2023Model, sun2009addressing}.

\section{Baseline Research}
\label{sec:Ontology}
\begin{wrapfigure}{r}{0.43\linewidth}
    \centering
    \vspace{-2.6mm}
    \includegraphics[width=1\linewidth]{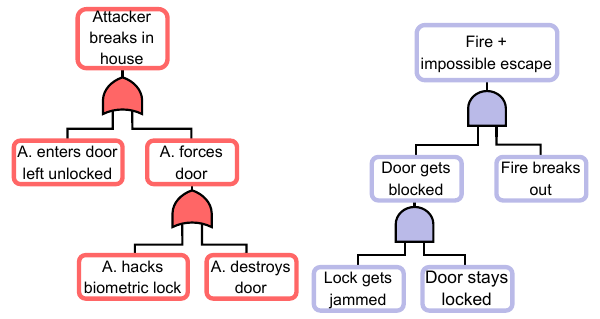}
    \caption{An attack tree (left) and a fault tree (right) for the locked door example.}
    \label{fig:AT_FT_Example}
\end{wrapfigure}
Fault Trees (\FTs) and Attack Tress (\ATs) constitute a sensible starting point as they are popular technologies that already encode key concepts highlighted in the \textit{Common Ontology of Value and Risk} (COVER) -- such as \textit{Events} and \textit{Actions} -- and pose a solid ground for model-based risk assessment. Fault tree analysis (FTA) \cite{RS15b} is a widespread technique to support safety risk assessment, and the use of fault trees is required, e.g., by the Federal Aviation Administration, the Nuclear Regulatory Commission, in the ISO 26262 standard \cite{I26262} for autonomous driving and for software development in aerospace systems. A \textit{fault tree} (\FT) (see \Cref{fig:AT_FT_Example}, right) models \textit{Events} that describe how component failures arise, and propagate disruption through the system, eventually leading to system-level failures. Leaves in a \FT represent \textit{basic events} (\BEs), i.e. elements of the tree that do not need further refinement. Once these fail, the failure is propagated through the \textit{intermediate events} (\IEs) via \textit{gates}, to eventually reach the \textit{top level event} (TLE), which symbolizes system failure. When considering model-based risk assessment of systems security -- \textit{attack trees} (\ATs) are widely employed. \ATs (see \Cref{fig:AT_FT_Example}, left) are hierarchical diagrams that represent malicious \textit{Actions} that can lead to a system being compromised \cite{Sch99, lopuhaa2022efficient}. \ATs are referred to by many system engineering frameworks, e.g.\ \emph{UMLsec} \cite{Jur02} and \emph{SysMLsec} \cite{RA15}, and are supported by industrial tools such as Isograph's \emph{AttackTree}~\cite{isoAT}. 
The \TLE of an \AT represents the attacker compromising the entire system, and the leaves represent \textit{basic attack steps} (\BASes): actions of the attacker that can no longer be refined. As for \FTs, intermediate nodes in \ATs are labelled with gates. 

The \textit{Common Ontology of Value and Risk} (COVER) \cite{sales2018common} 
is based on \textit{UFO} \cite{guizzardi2022ufo} -- a foundational ontology. It embeds a domain-independent conceptualization of risk, has been subject to validation and proper comparison to the literature of risk in risk analysis and management at large (e.g. \cite{iso73}), and it is built upon widespread definitions of risk. This ontology has already shown its utility in constructing formalisms for risk quantification and propagation \cite{fumagalli2023semantics}, and embeds several key assumptions about the nature of risk, which align with those in the literature in risk assessment. Firstly, risk is \textbf{experiential}. This means that the notions of ``event'' and ``object'' are deeply entangled and, when assessing the risk an object is exposed to, one aggregates risks ascribed to events that can impact the object. For instance, consider the risks your laptop is exposed to. To assess them, you will need to consider:
\begin{enumerate*}
    \item which of your goals depend on your laptop (e.g. work deliverables);
    \item what can happen to your laptop such that it would hinder its capability to achieve your goals (e.g. its screen breaking);
    \item which other events could cause these (e.g. you dropping it on the floor).
\end{enumerate*}
Then the risk your laptop (object) is exposed to is the aggregation of the risk of it falling (event) and breaking (event), and so on. The second assumption is that risk is \textbf{contextual}. Thus, the magnitude of the risk an object is exposed to may vary even if all its intrinsic properties (e.g., vulnerabilities or states) stay the same. To exemplify, let us pick one risk event involving your house door, namely that of robbers breaking into it. Naturally, the properties of the door -- such as having a strengthened blocking mechanism -- influence the magnitude of this risk. Still, the tools used by the robbers can significantly increase how risky the breaking event is. Lastly, another assumption that we derive from COVER is that risk is grounded on \textbf{uncertainty} about events and their outcomes. This is a very standard position -- see \cite{iso73} and \cite{aven2011ontological} -- which implies that likelihood is positively correlated with how risky an event is. For instance, the risk of encountering a bear is higher while walking in a forest than in an urban park, simply because it is more likely in the former case. 

In COVER, a \textsc{Risk Experience} is a multifaceted hypothetical occurrence that can be broken down into \textsc{Risk Events}, further categorized into \textsc{Threat Events} and \textsc{Loss Events}. \textsc{Threat Events} are hypothetical occurrences capable of precipitating \textsc{Loss Events}, which, in turn, are incidents that undermine the objectives of a \textsc{Risk Subject}, the \textsc{Agent} whose viewpoint is under scrutiny in the risk evaluation process. A \textsc{Loss Event} might involve \textsc{Objects at Risk} and \textsc{Risk Enablers}. 
Dispositions of objects at risk and risk enablers that can be manifested as threat and loss events are \textsc{Vulnerabilities}. As discussed in \cite{sales2018common} COVER accounts also for a numerical evaluation of \textsc{Risk} linked to a \textsc{Risk Assessment}. We emphasise here that the quantification of \textit{risk} is applicable solely to \textit{anticipated scenarios} that have the potential (though not certainty) to materialize. The ontology tackles this concern by acknowledging the feasibility of anticipated events, as discussed in \cite{DBLP:conf/er/Guarino17}.

 In order to ground $\OurFrame$ in COVER, we make a number design assumptions, which are used in the construction of both the new proposed model (\ODGs) and logic ($\OurLogic$). We further discuss how they relate to -- and are grounded in -- the COVER ontology, their implications and limitations. Throughout the paper, we highlight them with the \textbf{assumption} tag whenever they play an active role. We distinguish between two types of assumptions: \textit{operational} and \textit{structural}. Operational assumptions introduce constraints that are stricter than necessary due to the novel nature of this work, requiring a cautious and incremental approach. Structural assumptions, on the other hand, directly derive from the existing COVER ontology as-is. We elaborate on these in the sequel:

 \begin{itemize}
     \item[$\bullet$] \textbf{Assumption 1} (\textit{structural}): the attribution of impact on parent elements in the \ODG is independent of the attribution on children elements. COVER grounds this assumption by clearly distinguishing the notions of \textit{probability}, \textit{impact}, and \textit{risk}. On \ODGs we propagate disruption (i.e., attacks/failures propagation in the system) and -- quantitatively -- their probability values. In the case of disruption likelihood, the values assigned to each event depend on those assigned to the other events to which they are related. Differently, impact values are assigned independently to each single event. For example, the probability value assigned to a ``breaking laptop'' event naturally depends on the probability value of a possible related event such as ``stumbling''. This is not the case for what concerns the loss value (or associated impact) of the two events (in itself, stumbling may not be a problem, while laptop breaking represents something serious). Note that by distinguishing between probability and impact values, we enable a clearer assessment of an event's impact, independent of the likelihood of its occurrence; 
     \item[$\bullet$] \textbf{{Assumption 2}} (\textit{structural})\textbf{:} \OaRs that can participate in parent elements of an \ODG are a collection of all \OaRs that can participate in their children elements, plus additional \OaRs added by the risk assessor. This assumption is in line with the representation of events in COVER. In this context, events modelled as children of other events in the graph can be naturally taken as parts of the parent event, and in a simplified view, this allows inferring that objects that participate in parts (or sub-events) of an event, also participate in the event itself;
     \item[$\bullet$] \textbf{assumption 3} (\textit{operational})\textbf{:} for each \textit{\OaR} that participates in an element (event/action) of the \ODG, we assume that its parts participate in it as well, but the opposite does \textit{not} hold. For instance, if \textit{Door} participates in \textit{Door stays locked} also its part -- namely, \textit{Lock} -- participates in it, but if \textit{Lock} participates in \textit{Lock breaking} this does not imply that \textit{Door} participates in that event. This is, again, aligned with the theory about events, objects and their parts encoded by UFO and inherited by COVER\footnote{For a more detailed focus on this assumption we refer the reader to \cite{guizzardi2013towards,guarino2022events}.};
     \item[$\bullet$] \textbf{assumption 4} (\textit{operational})\textbf{:} in computing risk values, we assume the attacker already knows which failure occurred in the system before acting. Looking at the conceptualization provided in COVER, this aligns with the composition of the ``risk experience'' concept, which not only accounts for the role of ``passive'' elements involved in the assessment of risk (e.g., ``object at risk'') but also for the role of the ``active'' elements involved (see, for instance, the concept of ``threat object'' and related \textit{threat capability}). This, then, supports the two-step representation of the assessment process we propose, where, firstly, vulnerabilities in a system are identified and, secondly, based on these, threats can be activated. This solution allows for simulations that act as operationalizations of the concept of ``risk experience'';
     \item[$\bullet$] \textbf{assumption 5} (\textit{operational})\textbf{:} the attacker can adapt its strategy to the event under consideration. E.g., when computing max total risk (\Cref{sec:Logic}) we assume the attacker can maximise the risk level for each individual event in the graph by choosing the best actions at each iteration. Operating in this way gives practitioners the safest possible risk metric, as they are provided with a worst-case upper bound when computing total risk. Also here, the assumption is inspired by how the ``risk experience'' is represented in COVER, where the ``threat capability'' of a ``threat object'' participating in a ``threat event'' is always directly related to ``loss events'' and related risks.
 \end{itemize}
\section{Object-oriented Disruption Graphs}
\label{sec:ODGs}
\begin{figure}[h]
    \centering\includegraphics[width=.7\columnwidth]{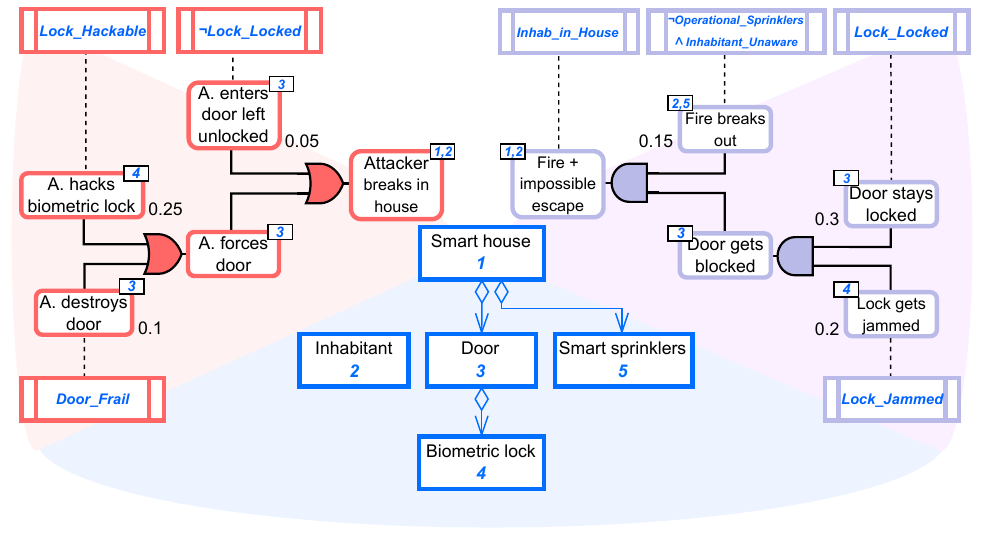}
    \vspace{-3mm}
    \caption{\ODG for the smart house locked door example.}
    \vspace{-1mm}
    \label{fig:ODG}
\end{figure}
\noindent Object-oriented DisruptiOn Graphs (\ODGs) extend classical \FT- and \AT-based risk analysis by integrating key concepts from COVER, i.e., adding needed constructs to reason about objects at risk during risk assessment. \Cref{fig:ODG} represents an object-oriented disruption graph that captures risks on a smart house. This scenario extends the well-known locked door example, typical of safety-security literature \cite{kriaa2014safety, Nicoletti2023Model, sun2009addressing}, with IS engineering specificities. On the left -- in \textcolor{red}{\contour{black}{\textbf{red}}} -- \textit{Actions} of an attacker are represented in an \AT. On the right -- in \textcolor{violet}{\contour{black}{\textbf{violet}}} -- \textit{Events} that can cause failures are represented in a \FT. Root nodes in both the \FT and \AT -- the \textit{top level events} (TLEs) -- can be mapped to COVER's \textit{Loss Events}. Each of the events/actions in the \FT and \AT is labelled with \textit{objects at risk} (\OaRs) that can \textit{participate} in it -- a white rectangle on the corner, containing numbers for objects that participate in the labelled event/action. These numbers refer to the \textit{Object Graph} -- at the bottom, in \textcolor{oars}{\contour{black}{\textbf{blue}}} -- where arrows in UML-like notation represent the \textit{parthood} relationship between objects. Specific combinations of properties of \OaRs that are needed for events/actions to happen -- what we call \textit{conditions} -- are typeset in blue and linked to events/actions via a dashed line. For example, the condition $\lnot\mathit{Lock\_Locked}$ is needed for the action \textit{Attacker enters door left unlocked} to happen. Finally, example probability values are typed next to each leaf node in the \FT/\AT components. With this model, one can ask ontology-aware questions that do not overlook the role of objects at risk, e.g., 
\begin{enumerate*}
    \item Given that an \textit{Attacker destroys the door} and that the \textit{Fire does not break out}, are any of the two loss events happening?
    \item Is the probability of both successfully forcing the door and fire breaking out lower than 0.05?
    \item What is the most risky event in which \textit{Inhabitant} participates, assuming that \textit{Lock is Locked}?
    \item What is the minimal risk level associated with the \OaR \textit{Door}, given all the events/actions in which it participates?
\end{enumerate*}
\noindent In \Cref{sec:DSL} we will formalize these queries via our logic ($\OurLogic$) and query language ($\OurDSL$).
\begin{definition}[\textit{Object-Oriented Disruption Graph}]
\label{def:ODG:syntax}
An \emph{Object-Oriented Disruption Graph} (\ODG) \odg is a tuple $(A, F, O, B)$ where $A$ is an attack tree, $F$ is a fault tree, $O$ is an object graph and $B$ is a disruption knowledge base.
\end{definition}
\vspace{-1.7mm}
\noindent As mentioned in \cite{stoelinga2021marriage}, \ATs and \FTs can be syntactically unified under the \textit{disruption tree} (\DT) model:
\begin{definition}[\textit{Disruption Tree}]
\label{def:Disruption_Trees}
A \emph{disruption tree} (DT) $\tree$ is a tuple $(N,E,t)$ where $(N,E)$ is a rooted directed acyclic graph, and $t\colon N {\rightarrow} \{\tOR,\tAND,\tLEAF\}$ is a function s.t. for $v \in N$, it holds that $t(v) {=} \tLEAF$ iff $v$ is a leaf.
Moreover, $\chOp\from$ $N\to 2^{N}$ gives the set of \emph{children} of a node and $\tree$ has a unique root, i.e., $R_\tree$.
\end{definition}
\vspace{-.7mm}
\noindent
We also define the set of intermediate events $\tIE = N\setminus \tLEAF$. Moreover, if \mbox{$u\in\child{v}$} then $u$ is called a \emph{child} of $v$, and $v$ is a \emph{parent} of $u$. 
Furthermore, we employ only \AND- and \OR-gates in the \AT/\FT components of the model. The behaviour of a \DT $\tree$ can be expressed through its \textit{structure function} \cite{RS15b} - $\sfunT$: if we assume the convention that a \LEAF has value 1 if disrupted and 0 if operational, the structure function indicates the status of the root node -- or \textit{top level event} (\TLE) -- given the status of all the \LEAFs of $\tree$. 
\noindent
Thus, for each set of \LEAFs we can identify its characteristic vector $\vec{b}$: we refer to this vector as a \textit{scenario}. We denote by $\allScenarios_\tree={2^{\LEAF_T}}$ the universe of scenarios of $\tree$. When further distinction is needed between \ATs and \FTs constructs, we use (respectively) the subscripts $\__{A}$ and $\__{F}$: e.g., we refer to a scenario on an \AT (resp. \FT) as an \textit{attack scenario} (resp. \textit{fault scenario}), represented by $\attack$ (resp. $\fault$). 
As shown before, in \FT- and \AT-related literature nodes canonically represent respectively \textit{events} and \textit{attack steps}: one might easily map \textit{attack steps} and \textit{events} to the terminology chosen in the COVER ontology, for which nodes $\NA$ of an \AT represent \emph{Actions} and nodes $\NF$ of a \FT represent \emph{Events}. From this point on, we will use the general term \textit{elements} to refer indistinctly to nodes in \FTs and \ATs.
\noindent To enrich \ATs and \FTs, we introduce \textit{Objects at Risk} (\OaRs) that explicitly capture impacted objects in (safety and security) risk experiences. 
\vspace{-.7mm}
\begin{definition}[\textit{Object Graph}]
\label{def:OG:syntax}
An \emph{object graph} (OG) $O$ is a rooted directed acyclic graph $(\NO,E_O,$ $OP, cOP)$ where:
\begin{enumerate*}
    \item nodes in $\NO$ represent \emph{Objects at Risk} (\OaRs); 
    \item directed edges in $E_O \subseteq \NO \times \NO$ represent the \emph{parthood} relation between \OaRs;
    \item \emph{properties} on \OaRs are atomic propositions $\op \in \emph{OP}$; and
    \item \label{def:OG:object_properties} $cOP \from \NO \to 2^{OP}$ 
    returns a set of atomic propositions of a node $v \in \NO$.
\end{enumerate*}
\end{definition}
\vspace{-.7mm}
\noindent Moreover, $\chOp\from$ $\NO\to 2^{\NO}$ gives the set of \emph{parts} of a node and \textit{O} has a unique root, denoted $R_O$. As previously hinted, \OaRs and the object graph (OG) are represented by connected blue rectangles (see \Cref{fig:ODG}, \cpageref{fig:ODG}). Similarly to \DTs' evaluation, we need a way to evaluate properties of \OaRs, thus:
\vspace{-.7mm}
\begin{definition}[\textit{Evaluating Properties of \OaRs}]
    We let a \emph{configuration} $\conf$ be the Boolean vector assigning values to properties of \OaRs in $OP$ and we let $\sfun_{\!O}\colon \mathbb{B}^n\times OP \to\mathbb{B}$ be a valuation such that $\sfun_{\!O}(\conf, op) = 1$ iff the Boolean value of a property $op \in OP$ equals $1$ given $\conf$.    
\end{definition}
\vspace{-.5mm}
\begin{wrapfigure}{R}{0.2\linewidth}
    \centering
    \includegraphics[width=1\linewidth]{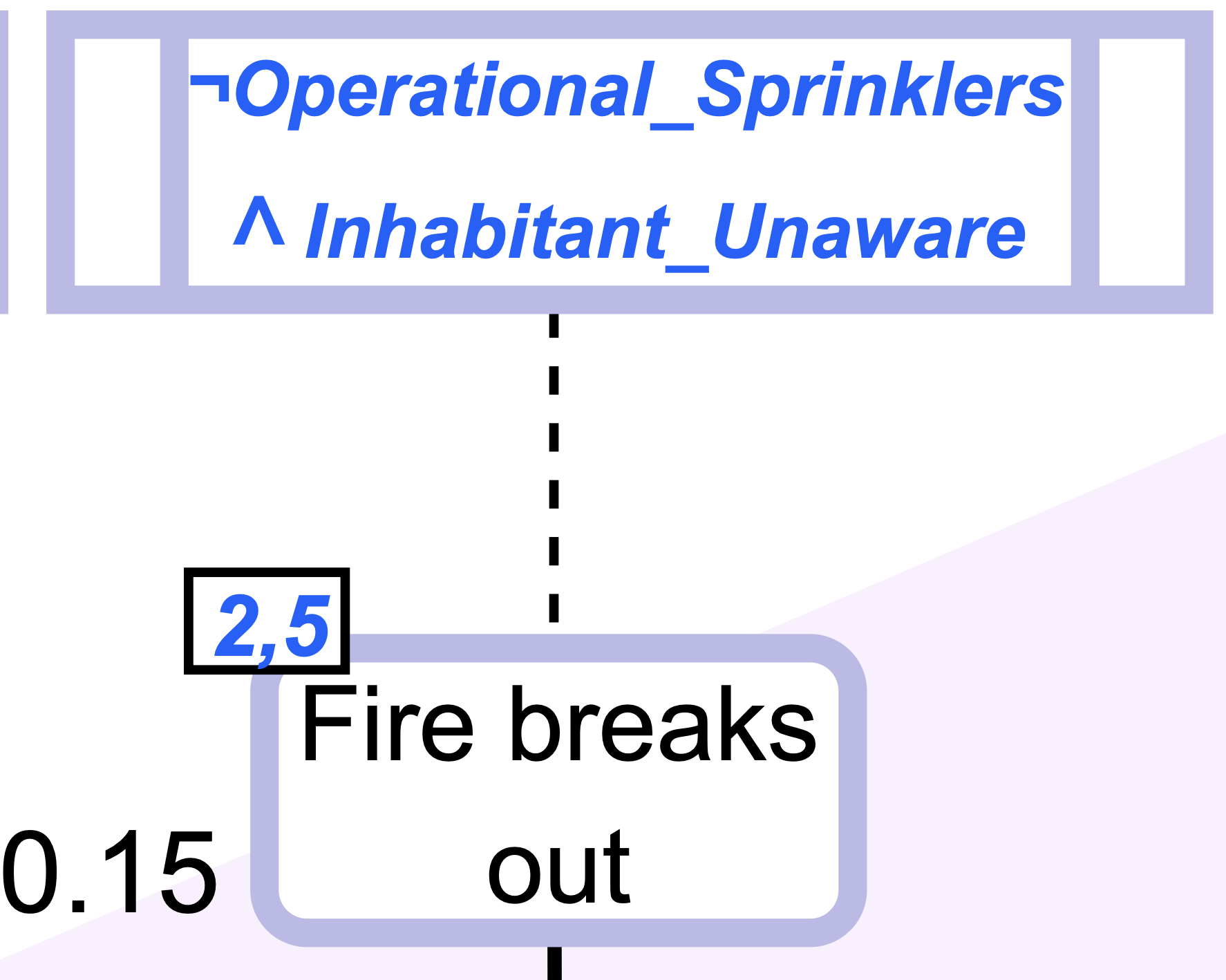}
    \caption{An excerpt of \Cref{fig:ODG}.}
    \label{fig:ODG_excerpt}
\end{wrapfigure}
\noindent Furthermore, we let $\allConfig$ be the set of all possible configurations. Finally, we introduce a \textit{Disruption Knowledge Base} (DKB) that establishes a formal relation between elements in \ATs and \FTs 
and \OaRs that can participate in them. 
We also define attribution of an impact value to \ATs and \FTs elements and their \textit{preconditions} and \textit{conditions} in the DKB. \textit{Preconditions} are relevant properties of \OaRs that can participate in a given event/action. These properties must be arranged in a specific way for events/actions to happen: \textit{conditions} express this arrangement.
\vspace{-2mm}
\begin{example}
\label{ex:properties_conditions}
Consider the excerpt of the \ODG of \Cref{fig:ODG} in \Cref{fig:ODG_excerpt}: conditions for the event \textit{Fire breaks out} -- in blue, connected with a dashed line -- are encoded in the Boolean formula $\lnot \mathit{Operational\_Sprinklers} \land \mathit{Inhabitant\_Unaware}$: in fact, the \OaRs \textit{Smart house} and \textit{Inhabitant} participate in the event \textit{Fire breaks out} and $\mathit{Operational\_Sprinklers}$ and $\mathit{Inhabitant\_Unaware}$ are the relevant properties of these participating objects at risk. These are exactly the \textit{preconditions} for event \textit{Fire breaks out} and they must be set resp. to \textit{false} and \textit{true} in conjunction for \textit{Fire breaks out} to happen. 
\end{example}
\vspace{-3mm}

\begin{definition}[\textit{Disruption Knowledge Base}]
\label{def:RB:syntax}
A \emph{disruption knowledge base} (DKB) $B$ is a tuple $(D, Im, Pa, Pr)$ where:
\begin{enumerate}
    \item $D = \NA \cup \NF \cup \NO$ is an \emph{entity domain} where $\NA, \NF$ and $\NO$ are pairwise disjoint 
    \item \label{def:RB:Impact} $Im \from \NA \cup \NF \to \mathbb{R}_{\geq 0}$ is a function that returns an \emph{impact factor} for each \emph{element} $v \in \NA \cup \NF$
    \item $Pa \from \NA \cup \NF \to 2^{\NO}$ is a function that for each \emph{element} $v \in \NA \cup \NF$ returns a set of \OaRs that can participate in $v$
    \item For each element $v \in \NA \cup \NF$, $Pr \from \NA \cup \NF \rightarrow 2^{OP}$ is a function that returns the set of its \emph{preconditions} $\mathit{Pr}(v) = cOP(o_1) \cup cOP(o_2) \ldots \cup cOP(o_n)$ with $o_i \in Pa(v)$
    \item \label{def:RB:Conditions} For each element $v \in \NA \cup \NF$, \emph{conditions} on $v$ are represented by a Boolean formula $\mathit{Cond}(v)$ over its \emph{preconditions} $Pr(v)$
\end{enumerate}
\end{definition}

\vspace{-1mm}
\noindent Do note that \textit{conditions} naturally map to \textit{descriptions of situations} in COVER. \textbf{Assumption 1:} Moreover, we assume that attribution of impact on parent elements (see \cref{def:RB:Impact}) is independent of the attribution on children elements. 
\textbf{Assumption 2:} We also assume that \OaRs that can participate in parent elements (events/actions) -- being decorated with AND-gates or OR-gates in the \DT -- are a collection of all \OaRs that can participate in their children elements, plus additional \OaRs added by the risk assessor (exactly what the $Pa$ function returns). 
As seen with properties of \OaRs, we need a way to evaluate whether conditions on a given \DT element (it being an action or an event) are satisfied:
\vspace{-.7mm}
\begin{definition}[\textit{Evaluating Conditions of Elements}]
\label{def:Evaluating_Conditions}
  We let $\sfun_{Cond}\colon \mathbb{B}^n\times Form \to\mathbb{B}$ be the evaluation function for conditions -- with $Form$ being the set of Boolean formulae -- such that given a configuration $\conf$ and conditions $Cond(v)$ for an element $v$, $\sfun_{Cond}(\conf, Cond(v)) = 1$ iff the Boolean assignment in $\conf$ satisfies $Cond(v)$.  
\end{definition}
\vspace{-.7mm}
\noindent To summarize: for each element $v$ we construct a Boolean formula $Cond(v)$ over the set of preconditions $\mathit{Pr}(v)$, which are exactly the relevant \textit{properties} of \OaRs that \textit{participate} in $v$ which are needed for $v$ to happen. To do so, we collect the $n$ \OaRs that can participate in $v$ with $Pa(v)$ and -- for each of these objects $o_i \in \{o_1, \ldots, o_n\}$ -- we collect its \textit{properties} via $cOP(o_i)$. \textbf{Assumption 3:} For each \textit{\OaR} that participates in element (event/action) $v$, we assume that its parts participate in $v$ as well, but the opposite does \textit{not} hold. 
\noindent
The union of all these collected sets $cOP(o_i)$ is exactly the set $\mathit{Pr}(v)$ of \textit{preconditions} on $v$. The Boolean formula $Cond(v)$ over preconditions in $\mathit{Pr}(v)$ for $v$ represents its \textit{conditions}. It is important to note that, in a practical setting, a user would iteratively be asked whether 
\begin{enumerate*}
    \item all participating objects in \textit{v} and
    \item all preconditions of a given participating \textit{\OaR}
\end{enumerate*}
 are relevant for conditions on \textit{v}. The ability to isolate and compute separately both preconditions and conditions will be functional in this setting. E.g., the property \textit{Lock\_Locked} is relevant for preconditions of an element  $v =\textit{Attacker enters door left unlocked}$, but \textit{Lock\_Hackable} is not: thus, it is not included as an atom in $Cond(v)$. As hinted, understanding computations on \ODGs requires further attention on the interplay between Actions (resp. Events) in \ATs (resp. \FTs) and their conditions. Since $\mathit{Cond}(v)$ is a Boolean formula over preconditions for $v \in \NA\text{ (resp. }v \in \NF)$, for $v$ to be attacked (to fail) the entire Boolean formula composed by $v \land \mathit{Cond}(v)$ must evaluate to \textit{true}. To account for this, let us define an extended structure function for \DTs where, for $v$ to be disrupted it would be necessary to have $\xfunT(\scenario, \conf, v)$ return 1, i.e., both the node in question should be disrupted and its conditions must be satisfied.
\vspace{-.7mm}
\begin{definition}[\textit{Extended Structure Function}]
\label{def:DT_Extended_Structure_Function}
The extended structure function of a disruption tree \tree is a function $\xfunT\colon \mathbb{B}^n\times \mathbb{B}^n\times N \to\mathbb{B}$ that takes as input a scenario $\scenario$, a configuration $\conf$ and an arbitrary element $v \in N$. We define it as follows:
\vspace{-4mm}
{\footnotesize  
  \setlength{\abovedisplayskip}{4pt}
  \setlength{\belowdisplayskip}{\abovedisplayskip}
  \setlength{\abovedisplayshortskip}{0pt}
  \setlength{\belowdisplayshortskip}{3pt}
\begin{align*}
\xfunT(\scenario, \conf, v) =&\begin{cases}
	b_i \land \sfun_{Cond}(\conf, Cond(v)) &
		\text{if}~v=v_i\in\tLEAF\\
	\underset{v'\in\mathit{\mathit{\Child}}(v)}{\bigvee} \xfunT(\scenario, \conf, v') \land \sfun_{Cond}(\conf, Cond(v))&
		\text{if}~v\in\tIE~\text{and}~t(v)=\tOR\\
	\underset{v'\in\mathit{\mathit{\Child}}(v)}{\bigwedge} \xfunT(\scenario, \conf, v') \land \sfun_{Cond}(\conf, Cond(v))&
		\text{if}~v\in\tIE~\text{and}~t(v)=\tAND
  \end{cases}
\end{align*}}  
\end{definition}
\vspace{-1.5mm}
\section{$\OurLogic$: a Logic to reason about \ODGs}
\label{sec:Logic}
We construct our logic on three syntactic layers, represented with $\lOne$, $\lTwo$ and $\lThree$. 
Layer 1 formulae reason about disruption propagation: the atomic propositions $a$ in $\OurLogic$ can represent any element in an \AT or \FT, and any property of \OaRs, i.e., $a \in N_A \cup N_F \cup OP$. Formulae can be combined through usual Boolean connectives. Furthermore, we can set evidence to construct what-if scenarios:  $\lOne[a \mapsto bool]$ sets the element $a$ in $\lOne$ to either 0 or 1, representing an event/action taking place or an object property being \textit{true} or \textit{false}. Finally, $\OurLogic$ allows reasoning about \textit{minimal risk scenarios} (\MRSs): minimal assignments on leaves of the \FT and \AT, such that a formula is satisfied (e.g., such that an event/action takes place). Note that \MRSs are always evaluated by fixing a specific configuration, i.e. a specific status of object properties. Layer 2 formulae reason about disruption propagation probabilities. We can check whether the disruption probability of a given $\lOne$ formula (e.g., a given event/action) is bounded by a selected threshold -- with our comparison operator $\bowtie\,\,\in \{<, \leq, =, \geq, >\}$ -- and we can also set probabilistic evidence, i.e., formulate scenarios where an event/action $e \in N_A \cup N_F$ is assigned a specific disruption probability $q$. Combining layer 2 formulae with Boolean operators is also allowed. Lastly, layer 3 reasons about safety- and security-related risk levels. One can ask what are the most risky actions/events in which an \OaR participates -- with $\ast \in \{{\color{red}A}, {\color{violet}F}\}$ symbolizing resp. \AT and \FT nodes, i.e., actions and events. Moreover, one can ask what is the max/min total risk level to which an \OaR is subject, aggregating risk from both safety- and security-related events/actions in which it participates, and what is the optimal configuration of object properties to minimise risk on an \OaR. Finally, one can also set evidence on object properties, i.e., forcefully set them to \textit{true} or \textit{false} to create insightful what-if scenarios. Note that when setting evidence we usually assign values to $a\in \LEAFs\cup OP$ in layer 1, and to $e\in \LEAFs$ in layer 2. We can however assign values to \IEs of \ATs and \FTs if
\begin{enumerate*}
    \item $a/e \in N_A \cup N_F$ is a module \cite{dutuit1996linear}, i.e., all paths between descendants of $a/e$ and the rest of the \AT or \FT pass through $a/e$
    \item and none of the descendants of $a/e$ are present in the formula. 
\end{enumerate*}
If so, we prune that (sub)\AT or (sub)\FT (and relative conditions) and treat occurring \IEs as \LEAFs. We can formally define the syntax for $\OurLogic$ as follows:
\vspace{-1mm}
{\footnotesize  
  \setlength{\abovedisplayskip}{6pt}
  \setlength{\belowdisplayskip}{\abovedisplayskip}
  \setlength{\abovedisplayshortskip}{0pt}
  \setlength{\belowdisplayshortskip}{3pt}
\begin{align*}
    &\text{\textcolor{gray}{Layer 1: }} \lOne &&\!\!\!\!\!\!\!\Coloneqq 
        a
        \mid \lnot \lOne 
        \mid \lOne \land \lOne
        \mid \lOne[a \mapsto bool] 
       \mid \opMRS(\lOne) \\
        &\text{\textcolor{gray}{Layer 2: }} \lTwo &&\!\!\!\!\!\!\!\Coloneqq
        \textsc{P}(\lOne) \bowtie \mathit{p} \mid
        \lnot\lTwo \mid
        \lTwo \land \lTwo \mid
        \lTwo[e \mapsto \mathit{q}]\\
        &\text{\textcolor{gray}{Layer 3: }}\, \lThree && \!\!\!\!\!\!\!\Coloneqq
        {\mathsf{MostRisky}_{\ast}(o)} \mid
        {\underset{\max}{\mathsf{TotalRisk}}(o)} \mid
        {\underset{\min}{\mathsf{TotalRisk}}(o)} \mid {\mathsf{OptimalConf}(o)} 
        \mid \lThree[op\mapsto bool]
\end{align*}}
\vspace{-9mm}
\section{Object-Oriented Risk Queries: $\OurLogic$ and $\OurDSL$}
\label{sec:DSL}
\input{tables/DSL_ODG}
To ease the usability of our logic, we present $\OurDSL$, a Domain Specific Language (DSL) for $\OurLogic$. Defining languages and tools to specify properties and requirements is common: in \cite{crapo2017requirements} the authors capture high-level requirements for a steam boiler system in a human-readable form with SADL. Further controlled natural languages for knowledge representation include Processable English (PENG) \cite{white2009update}, Controlled English to Logic Translation (CELT) \cite{pease2003english}, Computer Processable Language (CPL) \cite{clark2005acquiring} and FRETish \cite{conrad2022compositional}. $\OurDSL$ is constructed by adhering to the same design philosophy of $\mathit{LangPFL}$ -- a domain specific language for \FTs that was developed in the literature \cite{Nicoletti2023PFL}. As for $\mathit{LangPFL}$, $\OurDSL$ is inspired by the aforementioned languages for their ease of use and close proximity to natural language. $\OurDSL$ expresses only a fragment of $\OurLogic$. Notably, nesting of formulae is disallowed: we retain most of the expressiveness of $\OurLogic$ while making property specification easier. In $\OurDSL$, \ODG elements are referred to with their short label and each operator in $\OurLogic$ has a counterpart in the DSL: Boolean operators, $\mathsf{not}$, $\mathsf{and}$, $\mathsf{or}$, $\mathsf{impl}\ldots$; setting the value of \ODG elements to Boolean or probabilistic values, $\mathsf{set}$, $\mathsf{set\_prob}$; minimal risk scenarios MRSs
, $\mathsf{MRS[\ldots]}$
; operators to check disruption probability thresholds, $\mathsf{\mathsf{Prob}[\ldots]\bowtie\ldots}$ (note that $\bowtie \; \in \{<,\leq,=, \geq, > \}$); and to reason about risk levels aggregated on a given object and about risky actions/events in which this object participates, $\mathsf{MostRiskyA[\ldots]},$ $ \mathsf{MostRiskyF[\ldots]},$ $ \mathsf{MaxTotalRisk[\ldots]},$ $ \mathsf{MinTotalRisk[\ldots]},$ $ \mathsf{OptimalConf[\ldots]}$. One can specify properties in $\OurDSL$ by utilizing operators inside structured templates. Assumptions on the status of \ODG elements can be specified under the $\mathbf{assume}$ keyword. These assumptions will be automatically integrated with the translated formula accordingly, e.g., $\mathsf{set}$ or $\mathsf{set\_prob}$ will be translated with the according operators to set evidence, while other assumptions will be the antecedent of an implication. A second keyword separates specified formulae from the assumptions and dictates the desired result: $\mathbf{compute}$ and $\mathbf{computeall}$ compute and return desired values, i.e., probability values, and lists of events/actions/configurations and MRSs 
respectively, while $\mathbf{check}$ establishes if a specified property holds. 

In \Cref{tab:DSL} we exemplify some queries on the \ODG for the smart house locked door example in \Cref{fig:ODG}. These queries exemplify the expressive power enabled by \ODGs, $\OurLogic$ and $\OurDSL$ and are chosen to reflect the different Layers of our logic. It is important to note that syntactically the \ODG model does not represent answers to these queries right away, so one cannot read them from \Cref{fig:ODG} directly. Answers are computed as per semantics of both the model and the logic: e.g., the \textit{A. forces door} node is a composite element, and the computation of this complex probability value follows probability composition on the structure of the model in \Cref{fig:ODG}. 
\section{Enabling Risk Computations: $\OurLogic$ Semantics}
\label{subsec:semantics}
To enable object-oriented risk computations and to ground the meaning of formulae into the enriched model presented in \Cref{sec:ODGs}, we define formal semantics for $\OurLogic$. For the first layer of the logic, formulae are evaluated on the following model $\lOneModel = \langle \riskscenario, \conf, \odg \rangle$ where a \emph{risk scenario} $\riskscenario = (b_1, \ldots, b_k)$ is defined as $\riskscenario = \attack \cup \fault$, $\conf$ is a configuration and \odg is an \ODG. Formally:\\
{\vspace{-5mm}\footnotesize  
  \setlength{\abovedisplayskip}{6pt}
  \setlength{\belowdisplayskip}{\abovedisplayskip}
  \setlength{\abovedisplayshortskip}{0pt}
  \setlength{\belowdisplayshortskip}{3pt}
\begin{alignat*}{2}
  \lOneModel&\models a &&\text{iff }
  \begin{cases}
       \text{with } a \in \NA & \xfunA(\attack, \conf, a) = 1\\
       \text{with } a \in \NF & \xfunF(\fault, \conf, a) = 1\\
       \text{with } a \in OP & \sfun_{\!O}(\conf, a) = 1 
  \end{cases}\\ 
  \lOneModel&\models \lnot \lOne &&\text{iff } \lOneModel\not\models \lOne \\
  \lOneModel&\models \lOne \land \lOne'\,\, &&\text{iff } \lOneModel\models \lOne \text{ and } \lOneModel\models \lOne'\\
  \lOneModel&\models \lOne[a_{i} \mapsto bool] \,\,&&\text{iff }
  \begin{cases}
      \text{with } a_i \in \NA \cup \NF & \!\!\!\lOneModel' \models \lOne\text{ with } \riskscenario' = (b_1', \ldots, b_k') \in \lOneModel', \\& \!\!\! b'_i = bool \in \BB \text{ and } b'_j = b_j\text{ for } j \neq i\\
      \text{with } a_i \in OP & \!\!\! \lOneModel' \models \lOne\text{ with } \conf' = (b_1', \ldots, b_m') \in \lOneModel', \\& \!\!\! b'_i = bool \in \BB \text{ and } b'_j = b_j\text{ for } j \neq i
  \end{cases}\\
  {\lOneModel}&{\models \opMRS(\lOne)} && {\text{iff }\riskscenario \in \minSat{\lOne}}
\end{alignat*}}
{\color{black}With $\minSat{\lOne}$ we denote the $\textit{minimal satisfaction set}$ of risk scenarios for $\lOne$, i.e., the set of minimal risk scenarios $\riskscenario$ that satisfy $\lOne$ given $\odg$. We define $\minSat{\lOne}$ as follows: $\minSat{\lOne} = \{\riskscenario \mid \langle \riskscenario, \conf, \odg \rangle \models \lOne \land \nexists \riskscenario' . \riskscenario' \subseteq \riskscenario \land \langle \riskscenario', \conf, \odg \rangle \models \lOne\}$. Note that the set of all minimal risk scenarios for a given $\lOne$ -- i.e., $\minSat{\lOne}$ -- is always computed by fixing a specific configuration $\conf$ first.} 
\rmkStefano{Might be nice to have a proof stating that $\minSat{\lOne}$ is equivalent to the set of all \MCSs for $\lOne$ when $\lOne$ contains only \FT elements and no conditions, and likewise for minimal attacks and \ATs.}
\\ Layer two formulae require the introduction of probabilities. First, we need to decorate the leaves of the \AT and the \FT in $\odg$ with probability values. To do so, we let an \emph{attribution on $\odg$} be a map $\alpha\colon \LEAFs \rightarrow [0, 1]$. With a slight abuse of notation, we simply write $\alphaOdg$ for the probability attribution on both the leaves of the \AT $A$ and the \FT $F$ in $\odg$. We then let $\rho(\lOne)$ define the probability of a given layer one formula $\lOne$. Intuitively: given $\lOne$ and a configuration $\conf$, we consider every possible fault scenario $\fault \in \allScenarios_F$ on $F$ and how that would impact truth values of \FT nodes in $\lOne$. For each of these fault scenarios, we compute the maximal probability of successfully attacking \AT nodes in $\lOne$ under the given configuration $\conf$.  
\textbf{Assumption 4:} Note that -- with this setup -- we assume the attacker already knows which \FT nodes in $\lOne$ failed. 
Consequently, we let:
{$$
\footnotesize  
  \setlength{\abovedisplayskip}{3pt}
  \setlength{\belowdisplayskip}{\abovedisplayskip}
  \setlength{\abovedisplayshortskip}{0pt}
  \setlength{\belowdisplayshortskip}{3pt}
\problOne{\lOne, \conf} = \sum_{\fault \in \allScenarios_F} \mathsf{Prob}(\fault) \times \mathbb{P}_{A}(\mathsf{Set}(\lOne, \fault, \conf))
\vspace{-1mm}
$$}
where the probability associated to each fault scenario $\fault \in \allScenarios_F$ -- with $v$ as a \BE -- is calculated via $\mathsf{Prob}(\fault) = \prod_{i = 1}^{k} b_i \times \attr{v_i} + (1 - b_i) \times (1 - \attr{v_i})$ and where the maximal probability of successfully attacking $\lOne$ is given by multiplying attributions on \BASes in every minimal attack scenario for $\lOne$ -- $\attack \in \llbracket \lOne \rrbracket_{A}$ -- to then take the maximum between the resulting values of these attacks. Formally:
$$
\vspace{-2mm}
\footnotesize  
  \setlength{\abovedisplayskip}{3pt}
  \setlength{\belowdisplayskip}{6pt}
  \setlength{\abovedisplayshortskip}{0pt}
  \setlength{\belowdisplayshortskip}{3pt}
\mathbb{P}_{A}(\lOne) = \max_{\attack \in \llbracket \lOne \rrbracket_{A}} \prod_{v \in \attack} \attr{v}
$$
This last step is coherent with a more general framework for multiple metric computations on \ATs previously defined in \cite{lopuhaa2022efficient, nicoletti2023ATM}. Finally, we account for how every possible fault scenario would impact truth values of atomic propositions of \FT nodes in $\lOne$ by recursively defining $\mathsf{Set}(\lOne, \fault, \conf)$, with $\fault \in \allScenarios_F$:
{\footnotesize  
  \setlength{\abovedisplayskip}{6pt}
  \setlength{\belowdisplayskip}{\abovedisplayskip}
  \setlength{\abovedisplayshortskip}{0pt}
  \setlength{\belowdisplayshortskip}{3pt}
\begin{alignat*}{2}
  &\mathsf{Set}(a, \fault, \conf) &&=
  \begin{cases}
      \text{with } a \in \NA & a \\
      \text{with } a \in \NF & 
        \begin{cases}
            1 &\text{iff } \xfunF(a, \fault, \conf) = 1\\
            0 &\text{otherwise}
        \end{cases}\\
      \text{with } a \in OP & 
        \begin{cases}
            1 &\text{iff } \sfun_{\!O}(a, \conf) = 1\\
            0 &\text{otherwise}
        \end{cases}
  \end{cases}\\
  &\mathsf{Set}(\lnot \lOne, \fault, \conf) &&= \lnot\mathsf{Set}(\lOne, \fault, \conf) \\
  &\mathsf{Set}(\lOne \land \lOne', \fault, \conf) &&= \mathsf{Set}(\lOne, \fault, \conf) \land \mathsf{Set}(\lOne', \fault, \conf)\\
  &\mathsf{Set}(\lOne[a_{i} \mapsto bool], \fault, \conf) &&= \mathsf{Set}(\lOne, \fault, \conf)[a_{i} \mapsto bool]\\
  &{\mathsf{Set}(\opMRS(\lOne), \fault, \conf)} && = {\opMRS(\mathsf{Set}(\lOne, \fault, \conf))}
\end{alignat*}}
where $1$ and $0$ represent the \textit{true} and \textit{false} derived layer 1 formulae. Note that -- also due to $\mathsf{Set}$ -- some occurrences can lead to the application of the $\mathbb{P}_{A}$ function to either \textit{true} or \textit{false}, i.e., when $\lOne = 1$ or $\lOne = 0$. In these cases, we fix that $\mathbb{P}_{A}(1) = 1$ and $\mathbb{P}_{A}(0) = 0$. With $\bowtie\,\,\in \{<, \leq, =, \geq, >\}$ and an updated model $\lTwoModel = \langle \conf, \odg, \alphaOdg \rangle$, semantics for layer two formulae can be defined as follows:
{\footnotesize  
  \setlength{\abovedisplayskip}{4pt}
  \setlength{\belowdisplayskip}{\abovedisplayskip}
  \setlength{\abovedisplayshortskip}{0pt}
  \setlength{\belowdisplayshortskip}{3pt}
\begin{alignat*}{3}
 &\lTwoModel &&\models \textsc{P}(\lOne) \bowtie \mathit{p} &&\text{ iff } \problOne{\lOne, \conf} \bowtie p; \quad\quad \lTwoModel \models \lnot \lTwo \text{ iff } \lTwoModel \not\models \lTwo;\\
 &\lTwoModel &&\models \lTwo \land \lTwo' &&\text{ iff } \lTwoModel \models \lTwo \text{ and } \lTwoModel \models \lTwo';\\
 &\lTwoModel &&\models \lTwo[e_i \mapsto \mathit{q}] &&\text{ iff } \lTwoModel (\alphaOdg[\text{with}\attrOp(v_i) \mapsto \mathit{q}]) \models \lTwo \text{, with } v_i \in \NA \cup \NF 
\end{alignat*}
}
Note that the model $\lTwoModel$ for layer 2 formulae does not contain a risk scenario $\riskscenario$: this is because in computing probabilities we already account for both 
\begin{enumerate*}
    \item possible fault scenarios $\fault$ and 
    \item possible attack scenarios $\attack$.
\end{enumerate*}
Layer 3 formulae require further attention on \OaRs and the participation relation.
To compute the risk level associated with an \OaR $o$, given a certain configuration -- e.g., the risk level of \textit{Door}, given that $\mathit{Lock}\_\mathit{Locked}$ is set to \textit{false} -- we first identify the \emph{set of elements in which $o$ participates, and for which a satisfying risk scenario plus configuration exist}. We can consider events/actions as layer one atomic formulae $a$ s.t. $a \in \NA \cup \NF$, and -- with $\ast \in \{{\color{red}A}, {\color{violet}F}\}$ -- formally define this set as:
$$
\footnotesize  
  \setlength{\abovedisplayskip}{2pt}
  \setlength{\belowdisplayskip}{\abovedisplayskip}
  \setlength{\abovedisplayshortskip}{0pt}
  \setlength{\belowdisplayshortskip}{3pt}
\llparenthesis o \rrparenthesis_\ast {=} \left\{a \in N_{\!\ast} \mid o \in Pa(a) \land \exists \scenario_{\!\!\ast}, \conf .\xfun_{\!\!\!\!\ast}(\scenario_{\!\!\ast}, \conf, a) = 1\right\}
$$
Note that one might want to parameterize some elements of the given configuration $\conf$ to, e.g., compute optimal assignments to minimize risk on a given \OaR. To accommodate for this need, we let $\allConfigPrime$ be the set of configurations that could still be compatible with a partial Boolean assignment $[op\mapsto bool]$. E.g.:
\vspace{-1.5mm}
\begin{example}
\label{ex:Partial_Configurations}
    Assume we want to consider only the configurations compatible with setting the evidence that $Lock\_Locked$ is \textit{true}, i.e., $[Lock\_Locked\mapsto 1]$ and that we only have two other object properties to consider, the values of which are still not assigned. The resulting partial configuration can be represented as $\conf = (1, \cdot, \cdot)$, where $\cdot$ represents the unassigned values of remaining object properties. The set $\allConfig_{[Lock\_Locked\mapsto 1]}$ would then contain all possible configurations whose assignments are still compatible with $\conf = (1, \cdot, \cdot)$: e.g., $\conf' = (1, 1, 0)$ would be in $\allConfig_{[Lock\_Locked\mapsto 1]}$, while $\conf'' = (0, 1, 0)$ would be excluded from the set since the value of the first object property is set to zero (against $[Lock\_Locked\mapsto 1]$).
\end{example}
\vspace{-1.5mm}
\noindent We let $\partRisk = \sum_{a \in \llparenthesis o \rrparenthesis_{\color{red}A} \cup \llparenthesis o \rrparenthesis_{\color{violet}F}}(\rho(a, \conf)_{A, F} \times \mathit{Im}(a))$ represent the cumulative risk value on a specific \OaR $o$, given events and actions in which it participates. Intuitively, we sum the risk values from each event/action in which $o$ participates, resulting from the probability of each event/action times its impact factor. Given a set of configurations $\allConfig$, we let $\valFun$ define semantics for layer 3 formulae:
%
{\footnotesize  
  \setlength{\abovedisplayskip}{4pt}
  \setlength{\belowdisplayskip}{\abovedisplayskip}
  \setlength{\abovedisplayshortskip}{0pt}
  \setlength{\belowdisplayshortskip}{3pt}
\begin{alignat*}{4}
&\valFun\!\left(\mathsf{MostRisky}_{{\color{red}A}}(o)\right) &&= \argmax_{\substack{ \,a \in \llparenthesis o \rrparenthesis_{\color{red}A}}}  \max_{\substack{\conf \in \allConfig}}(\rho(a, \conf)_{A, F} {\times} \mathit{Im}(a)); &&&&\\
&\valFun\!\left(\mathsf{MostRisky}_{{\color{violet}F}}(o)\right) &&= \argmax_{\substack{ \,a \in \llparenthesis o \rrparenthesis_{\color{violet}F}}} \max_{\substack{\conf \in \allConfig }  }(\rho(a, \conf)_{A, F} {\times} \mathit{Im}(a)); &&&&\\
&\valFun\!\left({\underset{\max}{\mathsf{TotalRisk}}(o)}\right) &&=\max_{\substack{\conf \in \allConfig}} \partRisk; 
&&\!\!\!\!\!\!\!\!\!\!\!\!\!\!\!\!\!\!\!\!\!\!\!\!\!\!\!\!\!\!\!\valFun\!\left({\underset{\min}{\mathsf{TotalRisk}}(o)}\right) &&= \min_{\substack{\conf \in \allConfig}} \partRisk;\\
&\valFun\!\left(\mathsf{OptimalConf}(o)\right) &&=  \argmin_{\substack{\conf \in \allConfig}   } \partRisk;
&&\!\!\!\!\!\!\!\!\!\!\!\!\!\!\!\!\!\!\!\!\!\!\!\!\!\!\!\!\!\!\!\valFun\!\left(\lThree[op\mapsto bool]\right) &&= \mathsf{Val}_{\allConfig_{[op\mapsto bool]}}\!\left(\lThree\right)
\end{alignat*}}%
\noindent\textbf{Assumption 5:} Note that with semantics as given, the attacker can adapt its strategy to the node under consideration. E.g., when computing max total risk we assume the attacker can maximise the risk level for each individual node in the graph by choosing the best \BASes at each iteration. We then sum risk levels derived from each of these single-node worst-case scenarios. 
\section{Related work}
\label{sec:Related_work}
This paper directly relates to approaches that seek to combine \ATs and \FTs and increase their expressive capabilities. In this sense, numerous works attempt combinations of \FTs and \ATs into joint safety-security models: these are collected in a recent survey on model-based formalisms for safety-security risk assessment \cite{Nicoletti2023Model}. Of the 14 selected formalisms in \cite{Nicoletti2023Model}, 7 combine or extend \FTs and \ATs: Attack-Fault Trees \cite{ArnoldGKS15},  Component Fault Trees \cite{kaiser2003new}, Extended Fault Trees -- also known as Fault Trees with Attacks
 \cite{fovino2009integrating}, Boolean driven Markov processes (BDMPs)  \cite{bouissou2003new}, Attack Tree Bow Ties \cite{BT20}, Failure-Attack-CounTermeasure Graphs \cite{sabaliauskaite2015aligning}, and State/Event Fault Trees (SEFTs) \cite{roth:hal-00848640}. Of these formalisms, only BDMPs and SEFTs explicitly integrate properties of objects by joining \FTs and Petri nets, expressing that certain disruptions can only happen in certain states. However, both BDMPs and SEFTs do not explicitly address how 
 \begin{enumerate*}
     \item the \textit{parthood} relation between objects and 
     \item the \textit{participation} relation between objects and events/actions 
 \end{enumerate*}
can influence the propagation of disruptions and the computations of risk levels. Furthermore, they do not allow for aggregation of risk levels on a given object, nor do they allow to compute an optimal configuration of states to minimize risk. Lastly, \cite{iverson1990diagnosis} presents Object-Oriented Fault Trees (OFTs), where each \FT node is described by an object with instance variables containing information such as the node's parents, children and type. Despite the name, OFTs do not account for objects participating in different events represented by \FT nodes, nor can they account for risk aggregation on objects, given safety- and security-related events/actions. 
\section{Discussion}
\label{sec:Discussion}
To validate our approach, we intend to perform requirement elicitation from users: we then plan to translate these elicited requirements into concrete queries, modeled after those in \Cref{sec:DSL} that serve for now a simple exemplification purpose. The effectiveness of our approach will be evaluated on the basis of the types of queries we are able to capture. Additionally, we aim to conduct user studies via the development of a mock-up implementation to further evaluate the approach hereby presented. Ideally, we envision users to be risk analysts, as our approach is a natural extension of models that they might already be familiar with. Finally, as far as scalability is concerned, we expect to formulate novel symbolic model-checking algorithms that encode presented computationl semantics in Binary Decision Diagrams (BDDs) \cite{Bryant1990BDDs}: BDDs are promising as they have proven to be computationally successful and efficient already in the classical setting of FT and AT analysis \cite{BasgozeVKKS22}.

It is important to notice that in the case of \textit{operational} assumptions -- stricter than the absolute necessary -- we take care to never be incoherent with COVER: i.e., we could lift or weaken these assumptions in future work, while remaining still grounded in the COVER ontology. E.g., one could weaked \textbf{assumption 3} to account for the principle of mereological expansion or \textbf{assumption 4} -- fundamental for risk computation -- when considering a different ordering between failures and attacks, or again \textbf{assumption 5} to model different types of attackers. These diverse possibilities increase flexibility in modelling disruptive situations while remaining grounded in COVER.  

\section{Conclusion and future work}
\label{sec:Conclusion}
We presented $\OurFrame$, an ontology-aware framework for object-oriented risk assessment that exploits both the strengths of model-based formal methods and ontologies: by combining ontologies with probabilistic risk quantification models, we enriched the expressive power of \FTs and \ATs and presented a more expressive ontology-aware model (\ODGs), logic ($\OurLogic$) and a query language ($\OurDSL$). We chose COVER for its domain-independent nature and its foundation in a comprehensive analysis of existing work on risk ontologies. However, our approach remains flexible and does not preclude the integration of knowledge from other ontologies. Our research opens up interesting directions for future work. First, one could allow a more nuanced notion of propagation or of parthood, considering the parthood relationship of \OaRs and mereology. Furthermore, one could enrich \ODGs by introducing new concepts from (the COVER) ontology, e.g., the notion of \textit{goal}, or by introducing multiple \textit{risk assessors} viewpoints via multiple \ATs and \FTs components. Finally, one could consider the effect of weakening assumptions, as discussed in \Cref{sec:Discussion}.

%% file: tables/DSL_ODG.tex
\def\head#1{\larger[1.5]\sffamily\color{white}{#1}}
\begin{table*}[htb]
    \setlength{\belowcaptionskip}{-3pt}
    \centering
    \resizebox{0.999\textwidth}{!}{%
    \begin{tabular}{l|c|l}
    \rowcolor{black!85}\multicolumn{1}{|c}{\head{Natural Language}} & \multicolumn{1}{c}{\head{Property in $\OurLogic$}} & \multicolumn{1}{c|}{\head{$\OurDSL$}} \\ \hline
    \rowcolor{lightgray}
    $\begin{array}{cc}
     & \,\,\,\text{{Given that an \textit{Attacker destroys the door}}} \\
     & \,\,\,\text{{and that the \textit{Fire does not break out}, are }}\\
     & \,\,\,\text{{any of the two TLEs happening?}}
\end{array}$ &
$\begin{array}{cc}
& \mathit{TLE1} \lor \mathit{TLE2}\\
& [\mathit{ADD}\mapsto 1, \mathit{FBO} \mapsto 0]
\end{array}$ & 
$\begin{array} {ll}
    \textbf{{a}}&\hspace{-8.99mm} \textbf{{ssume:}} \\
    & \text{{set ADD = 1}}\\
    & \text{{set FBO = 0}}\\
     \textbf{{check:}} & \\
    & \text{{TLE1 or TLE2}}
\end{array}$ \\
$\begin{array}{cc}
     & \,\,\,\text{{What are all the \MRSs such that}} \\
     & \,\,\,\text{{both \textit{Loss Events} happen, given that \textit{Lock}}} \\
     & \,\,\,\text{{\textit{is Locked} and \textit{Door is Frail}?}}
\end{array}$ &  
$\begin{array}{cc}
     & \llbracket \mathit{TLE1}\land \mathit{TLE2} \\
     & [\mathit{LiL}\mapsto 1, \mathit{DiF}\mapsto1] \rrbracket_\lOneModel
\end{array}$ & 
$\begin{array} {ll}
    \textbf{{a}}&\hspace{-7.8mm} \textbf{{ssume:}} \\
    & \text{{set LiL = 1}}\\
    & \text{{set DiF = 1}}\\
    \textbf{{comp}}&\hspace{-1.0mm} \textbf{{uteall:}}\\
    & \text{{MRS[TLE1 and TLE2]}}\\
\end{array}$ \\
\rowcolor{lightgray}
$\begin{array}{cc}
     & \text{\,\,\,\,\,{Is the probability of both successfully}} \\
     & \text{\,\,\,\,\,{forcing the door and fire breaking out}} \\
     & \text{\,\,\,\,\,{lower than 0.05?}}
\end{array}$ &
$\begin{array}{cc}
     & \mathsf{Prob}(\mathit{AFD}\land\mathit{FBO})< 0.05
\end{array}$ & 
$\begin{array} {ll}
    \textbf{{a}}&\hspace{-8.99mm} \textbf{{ssume:}}  \\ \textbf{{check:}} & \\ & \text{{Prob[AFD and FBO]}} < \text{{0.05}}
\end{array}$ \\
$\begin{array}{cc}
     & \text{{What is the most risky event in which}} \\
     & \text{{\textit{Inhabitant} participates, assuming that \textit{Lock}}} \\
     & \text{{\textit{is Locked}?}}
\end{array}$ &
$\begin{array}{cc}
     & \mathsf{MostRisky}_{{\color{violet}F}}(Inhab.)[\mathit{LiL}\mapsto 1] 
\end{array}$
& 
$\begin{array} {ll}
    \textbf{{a}}&\hspace{-7.8mm} \textbf{{ssume:}} \\ 
    & \text{{set LiL = 1}}\\   \textbf{{comp}}&\hspace{-1.0mm} \textbf{{uteall:}}\\
    & \text{{MostRiskyF[Inhab.]}} 
\end{array}$ \\
\rowcolor{lightgray}
$\begin{array}{cc}
     & \text{\,\,\,\,\,{What is the max risk level associated}} \\
     & \text{\,\,\,\,\,{with the \OaR \textit{Door}, given all the}} \\
     & \text{\,\,\,\,\,{events/actions in which it participates?}}
\end{array}$
&
$\begin{array}{cc}
     & {\underset{\max}{\mathsf{TotalRisk}}(Door)}
\end{array}$
& 
$\begin{array} {ll}
    \textbf{{a}}&\hspace{-7.99mm} \textbf{{ssume:}} \\ 
\textbf{{comp}} &\hspace{-1mm}
    \textbf{{ute:}}\\
    & \text{{MaxTotalRisk[Door]}}
\end{array}$ \\
$\begin{array}{cc}
     & \text{{What is the minimum risk level on}} \\
     & \text{{\textit{Door}, assuming that \OaR \textit{Look} exhibits}} \\
     & \text{{property \textit{Lock Hackable}?}}
\end{array}$ &
$\begin{array}{cc}
     & {\underset{\min}{\mathsf{TotalRisk}}(Door)}[\mathit{LH}\mapsto 1]
\end{array}$
& 
$\begin{array} {ll}
    \textbf{{a}}&\hspace{-7.99mm} \textbf{{ssume:}} \\
    & \text{{set LH = 1}}\\ 
\textbf{{comp}} &\hspace{-1mm}
    \textbf{{ute:}}\\
    & \text{{MinTotalRisk[Door]}}
\end{array}$ \\
\rowcolor{lightgray}
$\begin{array}{cc}
     & \,\,\,\text{{What are the properties that all \OaRs must}} \\
     & \,\,\,\text{{exhibit, in order to minimise the risk}} \\
     & \,\,\,\text{{level associated with the object \textit{House},}} \\
     & \,\,\,\text{{assuming we fix that \textit{Door is Frail}?}} \\
\end{array}$ &
$\begin{array}{cc}
     & {\mathsf{OptimalConf}(House)}[DiF \mapsto 1]
\end{array}$
& 
$\begin{array} {ll}
    \textbf{{a}}&\hspace{-7.8mm} \textbf{{ssume:}} \\ 
    & \text{{set DiF = 1}}\\   \textbf{{comp}}&\hspace{-1.0mm} \textbf{{uteall:}}\\
    & \text{{OptimalConf[House]}}
\end{array}$ \\
\end{tabular}}
\caption{Risk queries for \odg  (\Cref{fig:ODG}) in natural language, $\OurLogic$ and $\OurDSL$.} \label{tab:DSL}
\end{table*}

%% file: main.bbl
\begin{thebibliography}{10}
\providecommand{\url}[1]{\texttt{#1}}
\providecommand{\urlprefix}{URL }
\providecommand{\doi}[1]{https://doi.org/#1}

\bibitem{BT20}
Abdo, H., Kaouk, M., Flaus, J.M., Masse, F.: {A new approach that considers cyber security within industrial risk analysis using a cyber bow-tie analysis} (2017)

\bibitem{ArnoldGKS15}
Arnold, F., Guck, D., Kumar, R., Stoelinga, M.: Sequential and parallel attack tree modelling. In: Proc. {SAFECOMP}. pp. 291--299 (2015)

\bibitem{aven2011ontological}
Aven, T., Renn, O., Rosa, E.A.: On the ontological status of the concept of risk. Saf. Sci.  \textbf{49}(8),  1074--1079 (2011)

\bibitem{BasgozeVKKS22}
Basg{\"{o}}ze, D., Volk, M., Katoen, J., Khan, S., Stoelinga, M.: {BDDs Strike Back - Efficient Analysis of Static and Dynamic Fault Trees}. In: {(NFM)}. vol. 13260, pp. 713--732 (2022)

\bibitem{bouissou2003new}
Bouissou, M., Bon, J.L.: A new formalism that combines advantages of fault-trees and markov models: Boolean logic driven markov processes. {RESS}  (2003)

\bibitem{Bryant1990BDDs}
Brace, K.S., Rudell, R.L., Bryant, R.E.: {Efficient implementation of a BDD package}. In: 27th ACM/IEEE Design Automation Conference. pp. 40--45 (1990)

\bibitem{clark2005acquiring}
Clark, P., Harrison, P., Jenkins, T., Thompson, J.A., Wojcik, R.H., et~al.: {Acquiring and using world knowledge using a restricted subset of English}. In: Flairs conference. pp. 506--511 (2005)

\bibitem{conrad2022compositional}
Conrad, E., Titolo, L., Giannakopoulou, D., Pressburger, T., Dutle, A.: {A compositional proof framework for FRETish requirements}. In: {CCP}. pp. 68--81 (2022)

\bibitem{crapo2017requirements}
Crapo, A., Moitra, A., McMillan, C., Russell, D.: {Requirements capture and analysis in ASSERT (TM)}. In: {RE}. pp. 283--291. IEEE (2017)

\bibitem{dutuit1996linear}
Dutuit, Y., Rauzy, A.: A linear-time algorithm to find modules of fault trees. IEEE Transactions on Reliability  \textbf{45}(3),  422--425 (1996)

\bibitem{fovino2009integrating}
Fovino, I.N., Masera, M., De~Cian, A.: Integrating cyber attacks within fault trees. Reliability Engineering \& System Safety  \textbf{94}(9),  1394--1402 (2009)

\bibitem{fumagalli2023semantics}
Fumagalli, M., Engelberg, G., Sales, T.P., Oliveira, {\'I}., Klein, D., Soffer, P., Baratella, R., Guizzardi, G.: On the semantics of risk propagation. In: RCIS (2023)

\bibitem{DBLP:conf/er/Guarino17}
Guarino, N.: On the semantics of ongoing and future occurrence identifiers. In: ER 2017. vol. 10650, pp. 477--490. Springer (2017)

\bibitem{guarino2022events}
Guarino, N., Baratella, R., Guizzardi, G.: Events, their names, and their synchronic structure. Applied Ontology  \textbf{17}(2),  249--283 (2022)

\bibitem{guizzardi2013towards}
Guizzardi, G., Wagner, G., de~Almeida~Falbo, R., Guizzardi, R.S., Almeida, J.P.A.: Towards ontological foundations for the conceptual modeling of events. In: ER. pp. 327--341. Springer (2013)

\bibitem{guizzardi2022ufo}
Guizzardi, G., et~al.: {UFO}: Unified foundational ontology. Appl. Ont.  \textbf{17}(1) (2022)

\bibitem{I26262}
{International Standardization Organization}: {ISO/DIS} 26262: Road vehicles, functional safety. \url{https://www.iso.org/standard/68383.html} (2018)

\bibitem{iso73}
{ISO}: {Risk Management - Vocabulary, ISO Guide 73:2009} (2009)

\bibitem{isoAT}
{Isograph}: {A}ttack{T}ree. \url{www.isograph.com/software/attacktree/} (Acc Mar 2023)

\bibitem{iverson1990diagnosis}
Iverson, D.L., Patterson-Hine, F.: A diagnosis system using object-oriented fault tree models. Proc. Artificial Intelligence for Space Applications pp. 341--9 (1990)

\bibitem{Jur02}
J{\"u}rjens, J.: {UMLsec}: Extending {UML} for secure systems development. In: {UML}\textasciitilde 2002 --- The Unified Modeling Language. vol.~2460, pp. 412--425 (2002)

\bibitem{kaiser2003new}
Kaiser, B., Liggesmeyer, P., M{\"a}ckel, O.: A new component concept for fault trees. In: {SCS}. pp. 37--46. Citeseer (2003)

\bibitem{kriaa2014safety}
Kriaa, S., Bouissou, M., Colin, F., Halgand, Y., Pietre-Cambacedes, L.: Safety and security interactions modeling using the {BDMP} formalism: case study of a pipeline. In: SAFECOMP. pp. 326--341. Springer (2014)

\bibitem{lopuhaa2022efficient}
Lopuha{\"a}-Zwakenberg, M., Budde, C.E., Stoelinga, M.: Efficient and generic algorithms for quantitative attack tree analysis. IEEE TDSC  (2022)

\bibitem{Nicoletti2023PFL}
Nicoletti, S.M., Lopuha{\"a}-Zwakenberg, M., Hahn, E.M., Stoelinga, M.: Pfl: A probabilistic logic for fault trees. In: FM 2023. pp. 199--221 (2023)

\bibitem{nicoletti2023ATM}
Nicoletti, S.M., Lopuha{\"a}-Zwakenberg, M., Hahn, E.M., Stoelinga, M.: {ATM: A Logic for Quantitative Security Properties on Attack Trees}. In: {SEFM} (2023)

\bibitem{Nicoletti2023Model}
Nicoletti, S.M., Peppelman, M., Kolb, C., Stoelinga, M.: Model-based joint analysis of safety and security: Survey and identification of gaps. Comp. Sci. Rev.  \textbf{50} (2023)

\bibitem{pease2003english}
Pease, A., Murray, W.: An english to logic translator for ontology-based knowledge representation languages. In: {NLP-KE}. pp. 777--783. IEEE (2003)

\bibitem{roth:hal-00848640}
Roth, M., Liggesmeyer, P.: {Modeling and Analysis of Safety-Critical Cyber Physical Systems using State/Event Fault Trees}. In: {SAFECOMP} (2013)

\bibitem{RA15}
Roudier, Y., Apvrille, L.: {SysML-Sec}: A model driven approach for designing safe and secure systems. In: {MODELSWARD}. pp. 655--664. {IEEE} (2015)

\bibitem{RS15b}
Ruijters, E., Stoelinga, M.: {F}ault {T}ree {A}nalysis: A survey of the state-of-the-art in modeling, analysis and tools. Comp. Sci. Rev.  \textbf{15--16},  29--62 (2015)

\bibitem{sabaliauskaite2015aligning}
Sabaliauskaite, G., Mathur, A.P.: Aligning cyber-physical system safety and security. In: Complex Systems Design \& Management Asia, pp. 41--53. Springer (2015)

\bibitem{sales2018common}
Sales, T.P., et~al.: The common ontology of value and risk. In: {ER} (2018)

\bibitem{Sch99}
Schneier, B.: Attack trees. Dr. Dobb’s journal  \textbf{24}(12),  21--29 (1999)

\bibitem{stoelinga2021marriage}
Stoelinga, M., Kolb, C., Nicoletti, S.M., Budde, C.E., Hahn, E.M.: The marriage between safety and cybersecurity: Still practicing. In: SPIN. pp. 3--21 (2021)

\bibitem{sun2009addressing}
Sun, M., Mohan, S., Sha, L., Gunter, C.: Addressing safety and security contradictions in cyber-physical systems. In: CPSSW. Citeseer (2009)

\bibitem{white2009update}
White, C., Schwitter, R.: {An update on PENG light}. In: {ALTA}. pp. 80--88 (2009)

\bibitem{zio2018future}
Zio, E.: The future of risk assessment. Reliability Engineering \& System Safety  \textbf{177},  176--190 (2018)

\end{thebibliography}
